\documentclass{article}

\usepackage{svg} 

\usepackage{PRIMEarxiv}

\usepackage[utf8]{inputenc} 

\usepackage{amsmath}
\usepackage[T1]{fontenc}    
\usepackage{hyperref}       
\usepackage{url}            
\usepackage{booktabs}       
\usepackage{amsfonts}       
\usepackage{nicefrac}       
\usepackage{microtype}      
\usepackage{lipsum}
\usepackage{fancyhdr}       
\usepackage{graphicx}       
\graphicspath{{media/}}     

\usepackage{booktabs} 
\usepackage{accents}  
\usepackage{multirow} 

\usepackage{algorithmic}
\usepackage{algorithm}
\usepackage{array}
\usepackage[caption=false,font=normalsize,labelfont=sf,textfont=sf]{subfig}
\usepackage{textcomp}
\usepackage{stfloats}
\usepackage{verbatim}

\usepackage{hyperref} 
\hypersetup{
    colorlinks=true,
    linkcolor=black,
    filecolor=magenta,   
    citecolor=blue,
    urlcolor=cyan,
    pdftitle={Overleaf Example},
    pdfpagemode=FullScreen,
    }

\usepackage{comment}

\title{Human-Robot Kinaesthetic Interaction Based on Free Energy Principle
}

\author{
  Hiroki Sawada, Wataru Ohata, Jun Tani$^*$ \\
  Cognitive Neurorobotics Research Unit \\
  Okinawa Institute of Science and Technology Graduate University \\
  Okinawa, Japan 904-0302.\\
  \texttt{\{hiroki.sawada1, wataru.ohata, jun.tani\}@oist.jp} \\
}

\begin{document}
\maketitle

\begin{abstract}
The current study investigated possible human-robot kinaesthetic interaction using a variational recurrent neural network model, called PV-RNN, which is based on the free energy principle.
Our prior robotic studies using PV-RNN showed that the nature of interactions between top-down expectation and bottom-up inference is strongly affected by a parameter, called the meta-prior, which regulates the complexity term in free energy.
The current study examines how changing the meta-prior $w$ in the interaction phase affects the counter force generated when an experimenter attempts to induce movement pattern transitions familiar to the robot through its prior training.
The study also compares the counter force generated when trained transitions are induced by a human experimenter and when untrained transitions are induced.
Our experimental results indicated that (1) the human experimenter needs more/less force to induce trained transitions when $w$ is set with larger/smaller values, (2) the human experimenter needs more force to act on the robot when he attempts to induce untrained as opposed to trained movement pattern transitions.
Our analysis of time development of essential variables and values in PV-RNN during bodily interaction clarified the mechanism by which gaps in actional intentions between the human experimenter and the robot can be manifested as reaction forces between them.

\end{abstract}


\section{Introduction}
Studies on social human-robot interaction have attracted considerable attention recently because of their practical applications, especially with using the linguistic modality \cite{guadarrama2013grounding, kanda2004interactive, wainer2014using}.
However, investigation of more direct interaction such as via kinaesthesia should be also indispensable when considering more embodied aspects or enactivism \cite{varela2017embodied} of human-robot interactions.
Although there have been a reasonable number of practical studies on human-robot interaction using kinaesthesia, which have contributed greatly to human-robot joint collaboration, human assistance, and user interface \cite{bahl2022human, kruger2009cooperation, peternel2017towards}, few studies have attempted to understand essential mechanisms underlying kinaesthetic interaction in light of embodied cognition, social cognition, and system-level neuroscience. 

In this regard, the current study investigated human-robot kinaesthetic interaction by applying system neuroscience theory, the free energy principle, proposed by Friston \cite{friston2005theory} which is consonant with enactivism \cite{ramstead2020free, bruineberg2018anticipating}.
Let us consider a situation in which a robot and a human dance, holding each other with both hands, executing memorized dance patterns.
If the robot initiates a particular pattern from memory with strong intention, the human counterpart might follow it without resisting because of the strong counter force.
On the other hand, if the robot generates a pattern without strong intention, the human counterpart might be able to shift to a different pattern without experiencing strong counter force.
Let us consider another situation. 
If the human counterpart attempts to induce a movement pattern that is familiar to the robot, such guidance should proceed easily without strong counter force, since the robot can infer the intended pattern immediately and can move as anticipated. On the other hand, if the human counterpart attempts to induce a movement pattern unfamiliar to the robot, such guidance should experience strong counter force, since the movement is neither inferential nor predictable for the robot.

We presume that these situations can be well explained by predictive coding \cite{friston2009predictive} and active inference \cite{friston2010action, friston2016active,parr2019generalised}, based on the free energy principle \cite{friston2005theory}.
Predictive coding provides a theory in which perception is achieved by inferring hidden causes for sensory observations that minimize the error between top-down prediction of sensation made by the generative model and the actual sensation.
On the other hand, active inference provides a theory for action generation in which an optimal action is inferred to minimize the error between the preferred sensation and the actual sensation resulting from the action.
These two are not separate, but integrated through a sensorimotor loop in embodied cognition in which a perceptual inference and action generation can be achieved simultaneously by minimising errors through iterative interaction between the top-down predictive/generative process and the bottom-up inference. 
Also in social embodied cognition, the dynamic interaction between the top-down pathway for predicting while acting on the others and the bottom-up pathway for inferring the actional intention of the counterpart through sensory observation should become a crucial element.
Accordingly, we presume that different interactions in the dance appear, depending on the balance between the strength of the top-down pathway and that of the bottom-up pathway.

Recently, the number of studies addressing application of free energy principle in cognitive robotics has increased \cite{ciria2021predictive,   lanillos2021active, taniguchi2023world}.
Maselli et al. \cite{maselli2022active} showed that the active inference model is able to characterise movements generated by the agent's intention to resolve multi-sensory conflict or to achieve an external goal, such as reaching with its arm to a certain point with the agent having a VR-vision of its arm that was tilted from the actual position to confuse the agent.
Tschantz et al. \cite{tschantz2020scaling} presented a working implementation of active inference to reinforcement learning that demonstrated efficient exploration and an order of magnitude higher sample efficiency in a high-dimensional task, such as a mountain-car environment.
Also, Pezzato et al. \cite{pezzato2023active}, showed that a robotic task, such as reactive action planning can be formulated as a free energy minimisation problem by introducing a hybrid combination of active inference and behaviour trees.
However, studies investigating human-robot interaction using the free energy principle remain few in number.

The author's group has conducted neurorobotic studies related to frameworks of predictive coding and active inference to develop various types of recurrent neural network (RNN) models \cite{tani2016exploring}.
However, an essential problem in these studies is that RNN models have difficulty in dealing with probabilistic properties hidden in interactions between robots and environments.
To tackle this problem, our group developed a probabilistic variational RNN model, called PV-RNN \cite{Reza2019}, based on the free energy principle.

An indispensable feature of PV-RNN is that free energy is computed as a sum of the negative accuracy term and the complexity term, weighted by a parameter $w$, called the meta-prior.
Here the complexity term represents the divergence between the approximated posterior probability distribution and the prior probability distribution in probabilistic latent variables allocated in PV-RNN.
More intuitively, the complexity term can be understood as the internal gap between top-down expectation and bottom-up sensory reality.
The model can learn to extract probabilistic structures hidden in data either by embedding them in nonlinear, deterministic dynamics of chaos by setting a large value of the meta-prior $w$ or into stochastic processes by setting a small $w$ \cite{Reza2019}.

By following the above study, Chame et al. \cite{chameICRA2020} conducted a human-robot bodily interaction experiment using a PV-RNN model and attempted to show possible effects of $w$ on the interaction dynamics.
However, this study is considered preliminary because only some snapshots of the experiments were shown and no rigorous analysis of the experimental results was made.
Wirkuttis et al. \cite{wirkuttis2021leading} performed simulation studies on synchronized imitative interaction of dyadic robots using a PV-RNN model in which robots observed the sensation of simplified vision and proprioception, but without kinaesthesis.
Through statistical analysis of repeated simulation experiments, these studies show that a robot with a larger $w$ in both the learning phase and the test phase tends to lead the counterpart robot set with smaller $w$ by projecting stronger actional intention to the counterpart. 

The current study examined human-robot kinaesthetic interaction by implementing a PV-RNN model in a real humanoid robot.
In particular, the study attempted to translate top-down intention or prior belief, formulated using the free energy principle to the force bodily exerted on the counterpart by conducting statistical analysis on data obtained from repeated experiments.
In this experiment, a humanoid robot controlled by a PV-RNN model is trained to generate movement patterns with specific transition probability distributions among them.
After the training phase, two types of human-robot kinaesthetic interaction experiments were conducted.
In the first experiment, while the robot was generating trained patterns, a human experimenter tried to physically induce a set of trained familiar movement pattern transition in the robot. 
We examined how the robot's rigidity in terms of the counter force in its response changed depending on $w$, which was set during the interaction phase.
The second experiment compared the counter force during responses in these two cases, i.e., when the experimenter induced the robot to proceed with learned movement pattern transitions and when the experimenter initiated previously unlearned pattern transitions.

Our intensive analysis of time-development of the latent variables, complexity term, and prediction error observed in these experiments under different conditions clarifies how bodily interaction between an experimenter and a robot proceeds and what sorts of neural activities are generated through top-down and bottom-up interactions during an experimental task.

\section{Model} 

\subsection{Free Energy Principle}
As mentioned in the Introduction, the current study used PV-RNN, based on the free energy principle, as the basic model.
The free energy principle assumes that brains learn as well as inferred for given observations, by minimizing free energy, defined in Eq. \ref{eq:F}.
\begin{equation}
\begin{aligned}\label{eq:F}
    \mathcal{F} = \underbrace{D_{\rm KL}[q_\phi({\boldsymbol{z}}|{\boldsymbol{X}})\Vert p_\theta(\boldsymbol{z})]}_{\rm complexity} - \underbrace{ \mathbb{E}_{q_\phi({\boldsymbol{z}}|{\boldsymbol{X}})}[\log p_\theta({\boldsymbol{X}}|{\boldsymbol{z}})]}_{\rm accuracy},
\end{aligned}
\end{equation}
$p_\theta({\boldsymbol{X}}|{\boldsymbol{z}})$ is the likelihood of the sensory observation $\boldsymbol{X}$, given the probabilistic latent variables $z$ which is parameterized by $\theta$.
$q_\phi({\boldsymbol{z}}|{\boldsymbol{X}})$ is the inference model parameterised by $\phi$.
As shown in Eq. \ref{eq:F}, free energy consists of two terms.
The first term indicates the $complexity$, which is the divergence between the approximate posterior probability distribution and prior probability distribution for the latent variable $z$, and the second term is the $accuracy$ in predicting the sensory observation.

\subsection{PV-RNN implementation}
PV-RNN is operated in two distinct phases, a training phase and an interaction phase.
In the training phase, we prepared a dataset (joint angle sequences of the robot) and trained the PV-RNN model using it. 
In the interaction phase, the robot and the human experimenter interact physically, such that the PV-RNN attempts to drive the robot's arms by predicting next-time-step target joint angles while it infers latent variables using actual joint angle readings.
Each operation phase is explained in detail below.

\subsubsection{Training Phase}
First, we start with computation during the training phase.
The evidence-free energy ${\mathcal{F}}^{train}$ as a loss function for training the PV-RNN with the joint angle observation from time step $1$ to $T$ is shown in Eq.\ref{eq:evidenceF} (the exact derivation should refer to \cite{ahmadi2019novel}).
\begin{equation}
    \label{eq:evidenceF}
\begin{split}
    {\mathcal F}^{train} &= \beta \times \underbrace{D_{\text{KL}} [ q_\phi(\mathbf{z}_1|\mathbf{d}_0,\mathbf{X}_1) \Vert p_\theta(\mathbf{z}_1)]}_{\rm complexity} \\
    +&  \underbrace{w \times \sum_{t=1}^{T-1} \mathbb{E}_{q_\phi(\mathbf{z}_{1:t}|\mathbf{d}_{t},\mathbf{X}_{t:T+1})}\big [D_{\text{KL}}
    [q_\phi(\mathbf{z}_{t+1}|\mathbf{d}_t,\mathbf{X}_{t+1:T+1})\Vert p_\theta(\mathbf{z}_{t+1}|\mathbf{d}_t)]\big]}_{\rm complexity}  \\
    -& \underbrace{\sum_{t=1}^T\mathbb{E}_{q_\phi(\mathbf{z}_{1:t-1}|\mathbf{d}_{t-1},\mathbf{X}_{t:T})}[\ln p_\theta(\mathbf{X}_t|\mathbf{d}_t)]}_{\rm accuracy},
\end{split}
\end{equation}
where $q(\phi)$ and $p(\theta)$ are the inference model parameterised by $\phi$ and the generative model parameterised by $\theta$, respectively.

As shown in Eq.\ref{eq:evidenceF}, the meta-prior $w$ weights the complexity term, except for the initial step. 
As described previously, the setting of meta-prior in the training phase strongly affects behavioural characteristics of the trained network.
Prior studies \cite{ahmadi2019novel, wirkuttis2021leading} showed that a network trained with a larger $w$ develops higher precision, i.e., smaller standard deviation, in the prior distribution, whereas a smaller $w$ develops lower precision in the prior.
More or less precision in the prior means stronger or weaker top-down belief, respecyively.
PV-RNN is composed of two variables $d, z$; in which the former is a deterministic latent variable, and the latter is a random latent variable sampled from a Gaussian distribution. 
During forward computation of PV-RNN, $d$ in layer $l$ ($l = 1$ is the bottom layer, which is closest to the output layer) at time step $t$ is computed as,
\begin{equation} \label{eq:hiddenlayer} 
\begin{aligned}
    \mathbf{h}^l_t = &\left(1 - \frac{1}{\tau^l}\right)\mathbf{h}^l_{t-1}
    + \frac{1}{\tau^l} \biggl( \mathbf{W}^{ll}_{dd}\mathbf{d}^l_{t-1} + \mathbf{W}^{ll}_{zd}\mathbf{z}^l_t
    + \mathbf{W}^{ll+1}_{dd}\mathbf{d}^{l+1}_{t} + \mathbf{b}_h \biggr), \\
    \mathbf{d}^l_t = &\tanh(\mathbf{h}^l_t).
\end{aligned}
\end{equation}
Here, $\mathbf{h}_t^l$ denotes the internal state of $\mathbf{d}_t^l$ before the activation function $\tanh$ is applied, and $\mathbf{b}_h$ denotes the bias term of $\mathbf{h}$.
$\tau^l$ is a time constant unique to each layer, which encourages the network to process information by following an intrinsic time scale of the layer \cite{yamashita2008, Pio-lopez-hierarchy-AIF2016, Schillaci2020, hwang2020dealing}.
Matrices $\mathbf{W}_{dd}$ and $\mathbf{W}_{zd}$ express intra- and inter-layer connectivity weights in the network, respectively.
At $t=1$, the input to $\mathbf{h}$ in the top layer is calculated only from $\mathbf{z}_1^l$ and $\mathbf{b}_h$. 

Each dimension of $\mathbf{z}_t^p$ is sampled over the prior distribution, which is a Gaussian distribution of mean $\pmb{\mu}_t^p$ and standard deviation $\pmb{\sigma}_t^p$ individually.
At $t=1$, $\mathbf{z}_1^p$ is sampled from a standard normal distribution, and $z_t^p$ for the following time steps is computed by,
%
\begin{equation} \label{eq:p} 
\begin{aligned}
    \pmb{\mu}^p_t &= \tanh(\mathbf{W}^{ll}_{d\mu}\mathbf{d}_{t-1} + \mathbf{b}_\mu^p),\\ 
    \pmb{\sigma}^p_t &= \exp(\mathbf{W}^{ll}_{d\sigma}\mathbf{d}_{t-1} + \mathbf{b}_\sigma^p),     
\end{aligned}
\end{equation}
\begin{equation}
    \mathbf{z}^p_t = {\pmb{\mu}^p_t} + {\pmb{\sigma}^p_t} * \pmb{\epsilon}_t, \quad \textrm{with }\pmb{\epsilon}_t\sim\mathcal{N}(\mathbf{0},\mathbf{I}),
\end{equation}
where $*$ indicates the element-wise product of the two vectors. 
This equation follows the idea of the conditional prior \cite{chung2015recurrent}.
Here, $\mathbf{W}^{ll}_{d\mu}, \mathbf{W}^{ll}_{d\sigma}, \mathbf{b}_\mu^p$ and $\mathbf{b}_\sigma^p$ are the weight matrices and the bias terms for $\pmb{\mu}_t^p$ and $\pmb{\sigma}_t^p$, respectively.
$\pmb{\epsilon}$ is a value sampled from the standard normal distribution. 
In order to make model parameters differentiable through the random latent variable, we use the reparametrization trick \cite{kingma2013auto} to sample $\mathbf{z}_t^p$.

On the other hand,  $\mathbf{z}_t^q$ in inference model $q_\phi$ is sampled from the approximate posterior probability distribution $q(\mathbf{z}_{t})$ which is a Gaussian distribution with mean $\mu_t^q$ and standard deviation $\sigma_t^q$ which are computed as,
%
\begin{equation} 
\begin{aligned}
    \pmb{\mu}^q_t &= \tanh( \mathbf{A}^\mu_t),\\
    \pmb{\sigma}^q_t &= \exp(\mathbf{A}^\sigma_t),\\
    \mathbf{z}^q_t &= \pmb{\mu}^q_t + \pmb{\sigma}^q_t * \pmb{\epsilon}_t, \quad \textrm{with }\pmb{\epsilon}_t\sim\mathcal{N}(\mathbf{0},\mathbf{I})
\end{aligned}
\end{equation}
$\mathbf{A}^\mu_t, \mathbf{A}^\sigma_t$ are adaptive variables which are optimised during training. 
During the training phase, $\mathbf{z}_t$ in Eq.\ref{eq:hiddenlayer} is sampled from the approximate posterior.
On the other hand, $\mathbf{z}_t^p$  is sampled from the prior distribution to compute the Kullback-Leibler divergence (KLD) between the approximate posterior and the prior, which is the complexity term of the loss function Eq.\ref{eq:evidenceF}.

The obtained $\mathbf{d}_t^1$ is used for computing the output layer, of which function is,
\begin{equation}\label{eq:output}
\mathbf{o}_t = \mathbf{W}_o \mathbf{d}_t^1 + \mathbf{b}_0,
\end{equation}
where $\mathbf{W}_o, \mathbf{b}_o$ is the weight matrix and the bias of the output layer, respectively. 

Finally, we compute the predicted output $\mathbf{\bar{x}}_t$.
Each dimension of the predicted output is represented in a probabilistic distribution using the Softmax of $N_{soft}$ elements.
Therefore, the $j$-th softmax element of the $i$-th dimension of the predicted output $\bar{x}_t^{i, j}$ is computed by,
\begin{equation}
    \bar{x}_{t, i, j} = \frac{\exp(o_{t, i, j})}{\sum_{j=1}^{N_{soft}} \exp(o_{t, i, j})}.
\end{equation}
As shown in Eq.\ref{eq:evidenceF}, the evidence-free energy $\mathcal{F}$, which is the loss function of this model, is the sum of the KLD between the approximate posterior $q(\mathbf{z}_t| \mathbf{x}_t)$ and prior $p(\mathbf{z}_t | \mathbf{d}_{t-1})$, and the negative log-likelihood calculated from the output $\mathbf{\bar{x}}_t$ and the perception $\mathbf{x}_t$.

At time step $t$, the KLD between the approximate posterior and prior is defined as,
\begin{equation}\label{eq:KLD_definition}
    D_{KL} [q(\mathbf{z}_t| \mathbf{x}_{t:T})||p(\mathbf{z}_t | \mathbf{d}_{t-1})]
    = \sum_i^{N_z} q(z_t^i| \mathbf{x}_{t:T}) \ln \frac{p(z_t^i | \mathbf{d}_{t-1})}{q(z_t^i| \mathbf{x}_{t:T})},
\end{equation}
where $z_t^i$ is the $i$-th dimension of $\mathbf{z}_t$, $T$ is the length of the sequence and $N_z$ is the dimension of $\mathbf{z}_t$ of each layer.
Here, $q(\mathbf{z}_t| \mathbf{x}_t)$ and $p(\mathbf{z}_t | \mathbf{d}_{t-1})$ are both assumed to follow multivariate Gaussian distributions with diagonal covariant matrices. Hence, they are expressed with the mean $\pmb{\mu}_t$ ($=[\mu_{t, 1}, \mu_{t, 2}, ..., \mu_{t, N_z}]$) and the standard deviation $\pmb{\sigma}_t$ ($=[\sigma_{t,1}, \sigma_{t,2}, ..., \sigma_{t, N_z]}$) as below,
\begin{equation}
    q(z_t^i| \mathbf{x}_{t:T}) = \frac{1}{\sqrt{2\pi(\sigma_{t, i}^q)^2}}\exp{\left[-\frac{1}{2} \left( \frac{z_t^i - \mu^q_{t, i}}{\sigma_{t, i}^q} \right) ^2\right]},
\end{equation}
\begin{equation}
    p(z_t^i| \mathbf{d}_{t-1}) = \frac{1}{\sqrt{2\pi(\sigma_{t, i}^p)^2}}\exp{\left[-\frac{1}{2} \left( \frac{z_t^i - \mu_{t, i}^p}{\sigma_{t, i}^p} \right) ^2\right]}.
\end{equation}
Therefore, the KLD loss $r_t$ between the approximate posterior and the prior is computed as,
\begin{equation}\label{eq:KLD}
r_t = \begin{cases}
{\sum_i D_{KL} [q(z_t^i| \mathbf{x}_t)||p(z_t^i | \mathcal{N}(0,1))]}, \quad \textrm{if } t=1,\\
{\sum_i D_{KL} [q(z_t^i| \mathbf{x}_t)||p(z_t^i | \mathbf{d}_{t-1})]}, \quad \textrm{otherwise}.
\end{cases}
\end{equation}
On the other hand, as an equivalent of the negative log-likelihood term of Eq.\ref{eq:evidenceF}, the KLD between the target and the predicted output is computed.
As mentioned above, the predicted output is represented as a probabilistic distribution $P(x_t^{i} | \mathbf{\theta})$ using Softmax.
The target sequence is also transformed into a probabilistic distribution $Q(x_{t}^{i})$ along the same line \cite{ahmadi2017can}.
Therefore, the KLD is computed as,
\begin{equation}
    \label{eq:negativeLog}
        e_t = \sum_i^{N_x} D_{KL} [Q(\mathbf{x}_{t, i}) || P(\mathbf{x}_{t, i})]
        = \sum_{i}^{N_x} \sum_j^{N_{soft}} Q(x_{t, i, j}) \ln \frac{Q(x_{t, i, j})}{P(x_{t, i} | \mathbf{\theta})}.
\end{equation}
The loss function of our model which is equivalent to the evidence-free energy $\mathcal{F}^{train}$ is computed as the weighted sum of Eq.\ref{eq:KLD} and Eq.\ref{eq:negativeLog}.
As shown in Eq.\ref{eq:evidenceF}, the KLD between the approximate posterior and prior is regularised with a meta-prior $\beta$ for $t = 1$ and $w$ for $t \neq 1$. Here, $\beta$ is the weighting parameter specifically for $t = 1$, which regulates the initial sensitivity of the PV-RNN model.
Therefore, the loss function of our model during the training phase is computed as,
\begin{equation} \label{eq:LossFunction}
    \mathcal{F} = \begin{cases}
    \sum_{t=1}^T \left( e_t + \frac{N_x}{N_z} \beta r_t \right), \quad \textrm{if } t=1, \\
    \sum_{t=1}^T \left( e_t + \frac{N_x}{N_z} w^t r_t \right), \quad \textrm{otherwise}
    \end{cases}
\end{equation}
where $N_x$ and $N_z$ are the $\Bar{\mathbf{x}}$ dimension and $\mathbf{z}$ dimension in each layer, respectively.
$T$ is the length of the dataset sequence.
The negative log-likelihood, which we denote as \emph{prediction error}, $e_t$ is proportional to $N_x$ and the KLD between the approximate posterior and prior is proportional to $N_z$, under the assumption that each $\Bar{\mathbf{x}}$ dimension and $\mathbf{z}$ dimension is independent.
Therefore, the meta-prior during the training phase $w^t$ is normalised by $N_x$ and $N_z$ so that it can be compared among PV-RNN models with different $\Bar{\mathbf{x}}$ dimension and $\mathbf{z}$ dimension.
Through training, weight matrices, biases and adaptive variables $\mathbf{A}_t^\mu, \mathbf{A}_t^\sigma$ are updated based on back-propagation through time (BPTT) \cite{werbos1990backpropagation, lillicrap2019backpropagation, rumelhart1986learning}.
\subsubsection{Interaction phase}
We performed a human-robot real-time interaction in the interaction phase where the pre-trained PV-RNN drives the robot’s movement by predicting joint angles of the next time step.
The forward computation part remains the same as that in the training phase.
However, there are some differences in the loss function used and in ways of updating variables.
Unlike the training phase, weight matrices and biases are not updated.
Only the adaptive variables $\mathbf{A}_t^\mu$ and $ \mathbf{A}_t^\sigma$ are updated so that the approximate posterior can adapt to the ongoing observation of the joint angle sequence.

\begin{figure}[htbp] %
\begin{center}
\includegraphics[width=0.60\columnwidth]{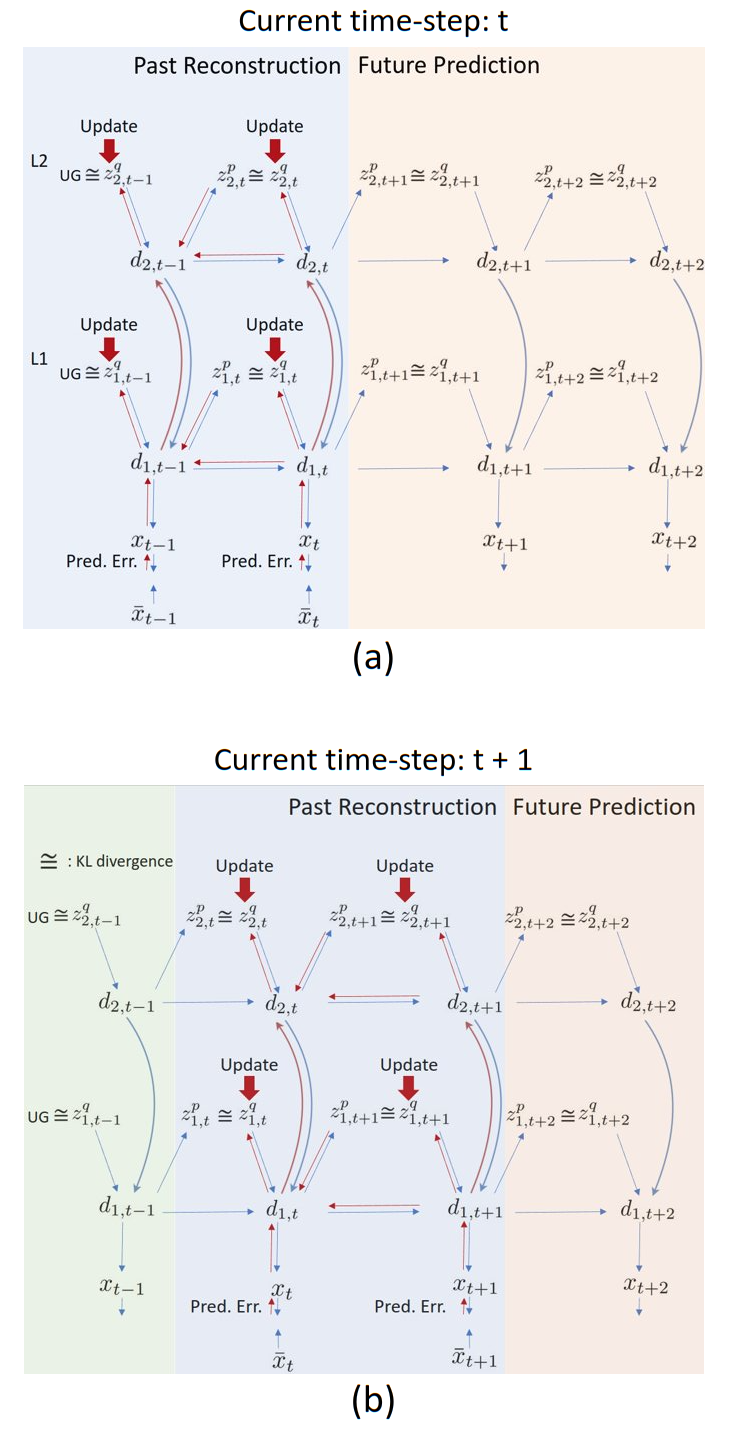}
\caption{Graphical representation of the interaction phase of PV-RNN. $d_{l, t}, z^p_{l, t}, z^q_{l, t}, x_{t}$ and $\Bar{x}_{t}$ indicates the deterministic latent variable, prior distribution, approximate posterior, output and target of layer $l$ and time-step $t$, respectively.
Blue and red arrows indicate forward propagation and backward propagation, respectively. The future, the past within the window and the past outside the window are coloured red, blue, and green respectively. The past window size is 2 in this graphical model. a) shows the network at time-step $t$. b) shows the network at time-step $t+1$.
}
\label{fig:PVRNN}
\end{center}
\end{figure}

PV-RNN predicts the future sensation and infers past latent variables using the past window spanning from time-step $t_{c}-t_{w}$ to the current time-step $t_c$ with a window size $t_w$ as shown in Fig. \ref{fig:PVRNN}.
The evidence-free energy in the interaction phase $\mathcal{F}^{int}$, normalized in the same way as Eq.\ref{eq:LossFunction}, is computed inside the past window as:
\begin{equation}
    \mathcal{F}^{int} = \begin{cases} 
    \sum_{t = t_c - t_w}^{t_c} \left( e_t + \frac{N_x}{N_z} \beta r_t \right), \quad \textrm{if } t = 1, \\
    \sum_{t = t_c - t_w}^{t_c} \left( e_t + \frac{N_x}{N_z} w^i r_t \right),
    \quad \textrm{otherwise},
    \end{cases}
    \label{eq:LossFunction_interaction}
\end{equation}
The adaptive variables $\mathbf{A}_t^\mu$ and $ \mathbf{A}_t^\sigma$ at each time-step $t$ from $t_{c}-t_{w}$ to $t_c$ are modified so as to minimize $\mathcal{F}^{int}$ by iterating the forward computation and the error back-propagation through time for a fixed number of times, called \emph{epochs}.
After modifying the approximate posterior by going through \emph{epochs} of iteration, next-step joint angles are predicted and fed into the PID controller of the robot in order to generate its movement.
Then, the past window is shifted one step ahead.
$w^i$, which is the meta-prior used in the interaction phase, could be set with different values from $w^t$ used in the training phase.
In \cite{ohata2020investigation}, a simulation experiment using a PV-RNN model shows that when a PV-RNN trained with a particular $w^t$ was reset with a smaller $w^i$ in the later interaction phase, the approximate posterior shifted away from the prior and the PV-RNN tended to adapt to the observed sensory sequence. 
This means that top-down intention in the PV-RNN became weaker.
On the other hand, when reset with a larger $w^i$, the PV-RNN tended to ignore the observed sensory sequence by generating its own intended patterns, which means that top-down intention becomes stronger.
However, the precision structure in the prior does not change even as $w^i$ is changed, since the prior distribution was developed in the learning phase.

\subsection{The employed robot controller}
\begin{figure*}[tbph]
  \centering
  \includegraphics[width=0.99\textwidth]{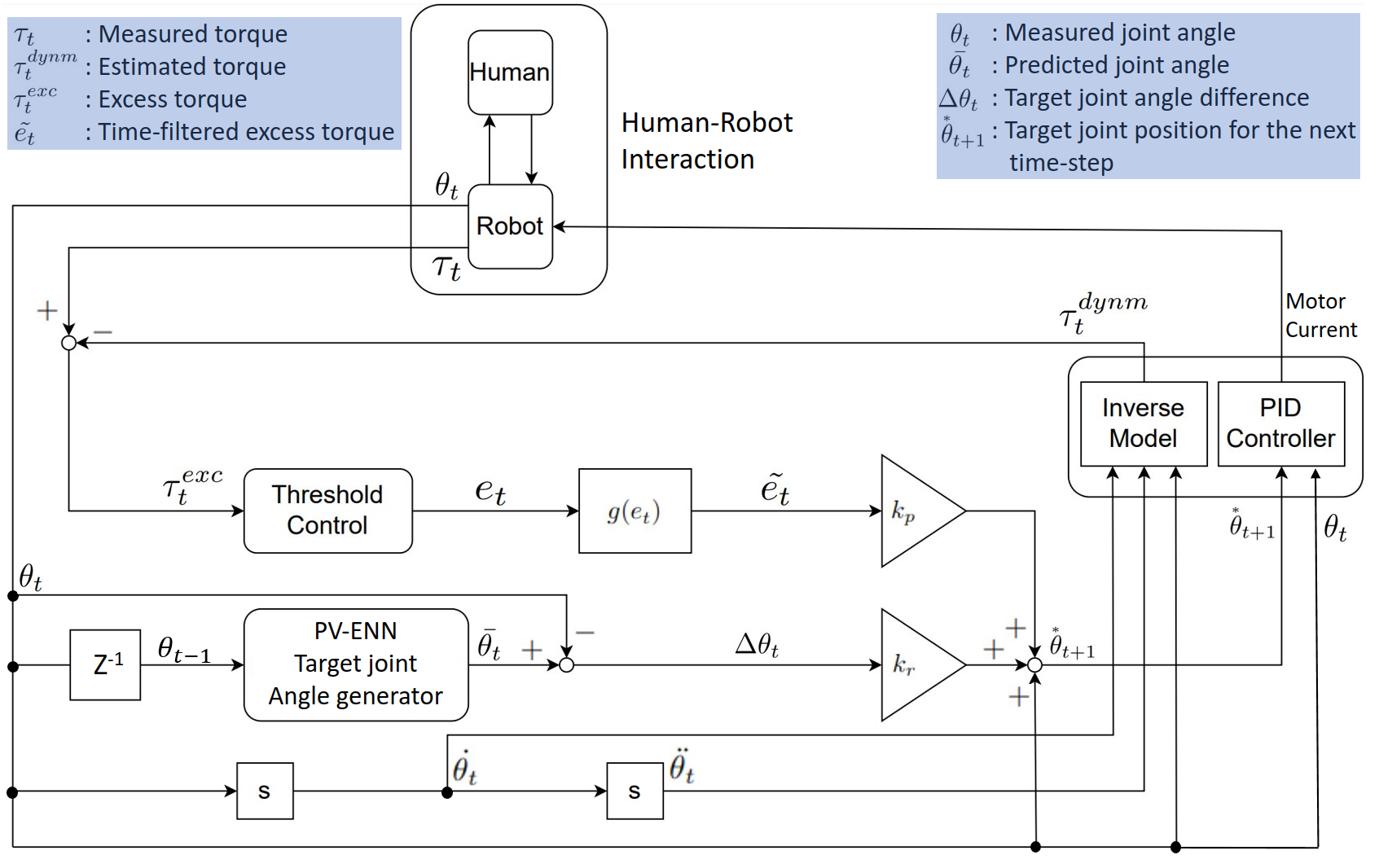}
  \caption{the human interacting with Torobo, the PV-RNN target joint angle generator, the inverse model, and the PID joint controller.}
  \label{fig:controller} 
\end{figure*}

The current study uses a humanoid robot, Torobo \footnote[1]{Humanoid-robot Torobo from Tokyo Robotics: \url{https://robotics.tokyo/products/torobo/}} to conduct human-robot kinesthetic interaction experiments.
Torobo is equipped with a built-in force-feedback controller that enables humans to back-drive joint angles with subtle force.
In the Torobo control system, when a human exerts certain torque on joints by pushing or pulling the limbs of Torobo, the exerted torque can be estimated by subtracting the torque inferred as necessary to account for the current static state as well as a dynamic state of the robot from the actual torque measured in the joints.
By computing the next time-step joint target positions by adding the current positions with the estimated exerted torque multiplied by a constant gain and feeding them in the PID controller, Torobo's limbs move by following the force exerted on them by the human.
This force-feedback controller is integrated with the PV-RNN, which generates the next time-step target joint angles (Fig.\ref{fig:controller}).

The overall control diagram is shown in Fig.\ref{fig:controller}, which includes a human experimenter interacting with Torobo, the PV-RNN target joint angle generator, the inverse model, and the PID joint controller.
The inverse model and PID controller were developed by the manufacturer of Torobo.
Details are as follows.
Excess torque, $\tau_t^{exc}$ which is torque exerted on joints by the human, can be estimated by extracting $\tau_t^{dynm}$ as the torque inferred for the current position, velocity, and acceleration of the joints by using the inverse model of Torobo from the current measured torque, $\tau_t$.
Next, excess torque, $\tau_t^{exc}$ is applied with a threshold control and $e_t$ as excess torque after threshold control is obtained.
Then, $e_t$ is time-filtered with the decay parameter $0< \alpha <1$, which generates $\tilde{e_t}$ as the time-filtered excess torque.
These operations are necessary to prevent unnecessary overreaction of the torque estimation against noise. 
These lines of processes are described in the following mathematical form:
\begin{equation} 
\begin{aligned}
\label{eq:ExcessTorque}
    \tau_t^{exc} &= \tau_t - \tau_t^{dynm}, \\
    e_t &= \begin{cases}
\max(0, \tau_t^{exc} - \tau^{th}), \quad \textrm{if } \tau_t^{ext} > 0,\\
\min(0, \tau_t^{exc} + \tau^{th}), \quad \textrm{otherwise},
\end{cases} \\
    \tilde{e_t} &=
    \begin{cases}
    e_t, \quad \textrm{for } t = 1,\\
    \max \Bigl( \alpha \times \tilde{e}_{t-1} + e_t, e_{max} \Bigl), \quad \textrm{otherwise},
    \end{cases}
\end{aligned}
\end{equation}
where $\tilde{e_t}$ has a fixed upper limit $e_{max}$.

The PV-RNN predicts $\bar{\theta_t}$ next time-step joint angles while performing the online inference, which is subtracted by $\theta_t$ as the current joint angle to generate $\Delta \theta_t$, the target joint angle difference shown as:
\begin{equation}
    \Delta \theta_t = \bar{\theta_t} - \theta_t
\end{equation}
Finally,  $\Delta \theta_t$ representing the next move intended by the PV-RNN and $\tilde{e}_t$ representing the estimate of torque exerted by the human are multiplied by each gain $k_r$ and $k_p$, respectively and they are added to the current joint angles to generate the final target joint positions, $\bar{\theta_t}$, which are fed into the PID controller as shown in the following:
\begin{equation}
\label{eq:target joint angle}
    \accentset{\ast}{\mathbf{\theta}}_{t+1} = \Delta \theta_t \times k_r + \tilde{e}_t \times k_p + \theta_t,
\end{equation}

\section{Experiment}

\subsection{Experiment Setup}

\subsubsection{Train Data-Set Preparation}
\begin{figure}[htbp]
  \centering
  \includegraphics[width=0.5\textwidth]{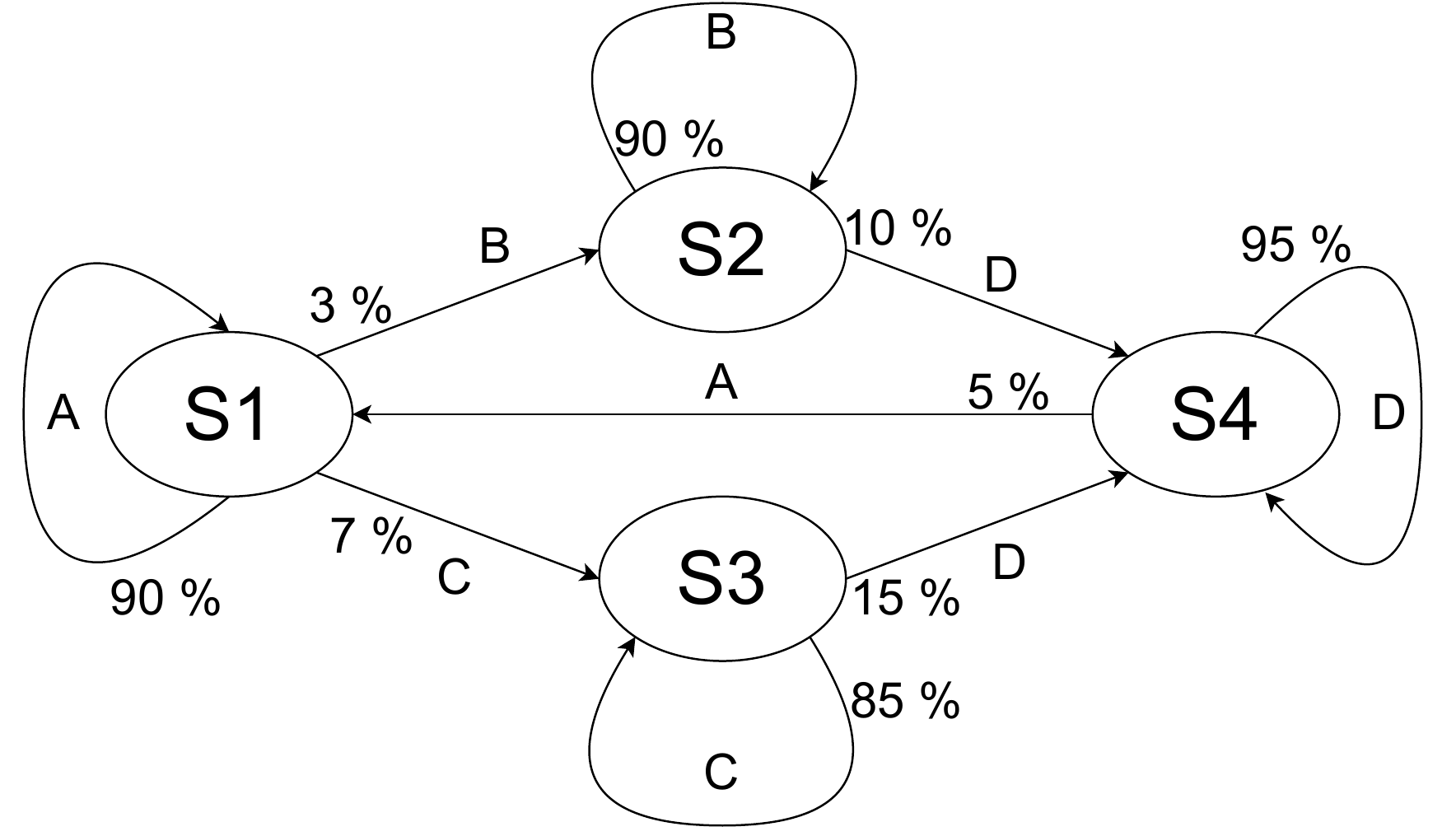}
  \caption{Schematic of the probabilistic finite state machine from which training data was generated.}
  \label{fig:PFSM}
\end{figure}

The proposed PV-RNN model was trained in a supervised manner by preparing 4-dimensional joint angle teaching trajectories.
In preparation for teaching trajectories, we considered four types of cyclic movement patterns (20 time-steps for each cycle) using Torobo's shoulder and elbow joint angles in both arms.
Then, it was assumed that cycling movement patterns transit from one to another following a probabilistic finite state machine (Fig.\ref{fig:PFSM}).
For example, after a movement pattern \emph{A} is generated for one cycle with the state at \emph{S1}, \emph{A} can be generated for one more cycle with $90\%$ probability staying at the same state, or \emph{B} or \emph{C} can transit to \emph{S2} or \emph{S3} with a probability of $3\%$ and $7\%$, respectively. 
We prepared 10 sequences that each consisted of 200 cycles of movement patterns, extending 4000 time-steps.

\subsubsection{The Network Configuration and Training}
\begin{figure}[htbp]
  \centering
  \includegraphics[width=0.7\textwidth]{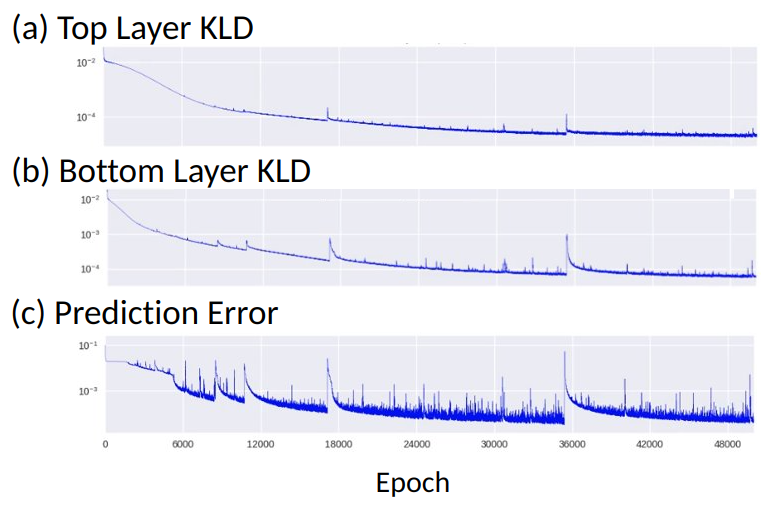}
  \caption{
    Resultant time-development from the training phase of one of the PV-RNN models.
  Development of KLD with respect to the number of training epochs is shown for (a) the Top layer (layer 2) and (b) the Bottom layer (layer 1).
  Development of the average prediction error over time steps in each teaching sequence is shown in (c).}
  \label{fig:train_result}
\end{figure}

To conduct human-robot dyadic interaction experiments, PV-RNN was trained 3 times with identical parameters (Table.\ref{tab:parameters}).
$\#\mathbf{d}, \#\mathbf{z}, \tau, \mathbf{w}^t, \mathbf{w}^i$ indicates the number of d neurons and z neurons, time constant, and meta-prior during the training phase and interaction phase, respectively.
The network was trained for 50,000 epochs to minimize the evidence-free energy shown in Eq.\ref{eq:evidenceF}, starting with random weights generated with different seeds for each training.
\begin{table}[htbp]\centering
\caption{PV-RNN Parameters}
\label{tab:parameters}
\begin{tabular}{cccccc}
\toprule\footnotesize
     & $\#\mathbf{d}$ & $\#\mathbf{z}$ & $\tau$ & $\mathbf{w}^{t}$ & $\mathbf{w}^{i}$\\
\cmidrule(lr){2-6}
\textbf{Layer 1 (Top)} & 60 & 6 & $3$  & 0.01 & $[0.01, 0.05, 0.1]$\\
\textbf{Layer 2 (Bottom)} & 30 & 3 & $9$ & 0.01 & $[0.01, 0.05, 0.1]$ \\
\bottomrule
\end{tabular}
\end{table}
%
Fig.\ref{fig:train_result} shows one of the resultant training processes.
It can be seen that prediction error and the KL divergence in both the top and the bottom layers decreased throughout training.
All three training processes converged in a similar way, achieving prediction errors and KL-divergences shown in Table.\ref{tab:TrainingResult}.

\begin{table}[htbp]
\centering
\caption{Training processes of all three PV-RNN model}
\label{tab:TrainingResult}
\begin{tabular}{ ccc  c c }
\cmidrule(lr){1-5}
 && \multicolumn{3}{c}{Epoch} \\
\cmidrule(lr){3-5}
\cmidrule(lr){3-5}
     & Model &    2,000     &       5,000       &    50,000\\
\cmidrule(lr){2-5}
\multirow{3}{*}{KL divergence} & 1 & $2.0 \times 10^{-3}$ & $5.4 \times 10^{-4}$ & $4.4 \times 10^{-5}$\\
& 2     &  $2.4 \times 10^{-3}$ &  $9.1 \times 10^{-4}$  & $5.6 \times 10^{-5}$\\
& 3    & $2.5 \times 10^{-3}$ &  $7.2 \times 10^{-4}$  &  $6.9 \times 10^{-5}$\\
\cmidrule(lr){1-5}
\multirow{3}{*}{Pred. Err.} &  1       &$1.8 \times 10^{-2}$ &  $7.2 \times 10^{-4}$  & $1.4 \times 10^{-4}$\\
& 2     & $1.4 \times 10^{-2}$ &  $6.6 \times 10^{-3}$  & $4.6 \times 10^{-5}$\\
& 3    & $1.3 \times 10^{-2}$  &  $1.6 \times 10^{-3}$  &  $1.4 \times 10^{-4}$\\
 \hline
\end{tabular}
\end{table}
%

\subsubsection{Evaluation of the trained networks}
Trained PV-RNNs were evaluated on the basis of how closely the state transition probability wasreconstructed compared with the target (Fig.\ref{fig:PFSM}). 
For this purpose, we conducted \emph{Prior Generation} in which the forward computation by following Eq.\ref{eq:hiddenlayer}-\ref{eq:output} was performed without any external observation for 40,000 time-steps for each trained network.
Then, probabilities for all possible movement pattern transitions were measured during prior generation and the resulting state transition probability was inferred (Table.\ref{tab:TrainEvaluation}).
The state transition probability computed for all trained networks is quite similar to the target one. 
Output and network dynamics of movement pattern transitions during prior generation are shown in Appendix, Fig.\ref{fig:PriorGeneration}.
\begin{table}[htbp]
    \centering
    \caption{Transition probabilities of the \emph{Prior Generation} and data set}
    \begin{tabular}{cccccc}
    \cmidrule(lr){1-6}
    Model & \emph{S1} $\rightarrow$ \emph{S2} & \emph{S1} $\rightarrow$  \emph{S3} & \emph{S2} $\rightarrow$  \emph{S4} & \emph{S3} $\rightarrow$  \emph{S4} & \emph{S4} $\rightarrow$  \emph{S1} \\
    \cmidrule(lr){2-6}
    1 & 2.8 \% & 3.5 \% & 15.6 \% & 12.5 \% & 3.1 \% \\
    2 & 2.7 \% & 9.3 \% & 14.1 \% & 11.5 \% & 7.2 \% \\
    3 & 3.1 \% & 4.1 \% & 9.8 \%  & 12.3 \% & 5.1 \% \\
    Target & 3 \% & 7 \% & 10 \% & 15 \% & 5 \% \\
    \cmidrule(lr){1-6}
    \end{tabular}
    \label{tab:TrainEvaluation}
\end{table}

\subsubsection{Human-Robot Interaction}
After the training phase, we conducted two types of human-robot interaction experiments using trained PV-RNNs.
In these experiments, while Torobo was generating movement pattern transitions successively based on the training, the human experimenter attempted to induce various movement pattern transitions by grasping both arms of Torobo and exerting force on them. 
These movement pattern transitions included trained transitions (\emph{AB, AC, BD, CD, DA}) and untrained transitions (\emph{AD, DB, DC, BA, CA}) in which \emph{AB}, for example, dictates that \emph{A} pattern is forced to transit to \emph{B} pattern.
Each transition from one pattern to another requires some guiding force by a human experimenter, even for trained transitions, since ongoing patterns tend to repeat another cycle with high probability, more than $85\%$ for all patterns.
This means that there exist some conflicts between movement trajectories intended by the robot and those by the human experimenter during both trained and untrained transitions.

Experiment-1 examined the effect of $w^i$ settings on the interactions.
While Torobo was generating movement pattern transitions successively for 2,200 time-steps, the human experimenter attempted to induce one of the trained transition every 200 time-steps starting from $t = 200$ which resulted in 10 trained transitions.
The duration of each attempt lasted 100 time-steps at the most.
This experiment was conducted three times for all three trained PV-RNNs by changing $w^i$ with $0.01$, $0.05$, and $0.1$.

Experiment-2 examined the difference between trained and untrained transitions by setting $w^i$ with a fixed value.
The experimental procedure was the same as in Experiment-1, although this time, $w^i$ was fixed at $0.01$ for both layers.
This experiment, using all three trained PV-RNNs, resulted in 30 untrained transition attempts.
In these experiments, we recorded the time development of essential values including latent variables, predicted and observed joint angles, prediction error, and the KL-divergence in both layers for later analysis of experiment results.

\subsection{Experiment Results}
In this section, we show the results of the aforementioned experiments.

\subsubsection{Experiment-1}
\begin{figure*}[h] %
\begin{center}
\includegraphics[width=0.95\linewidth]{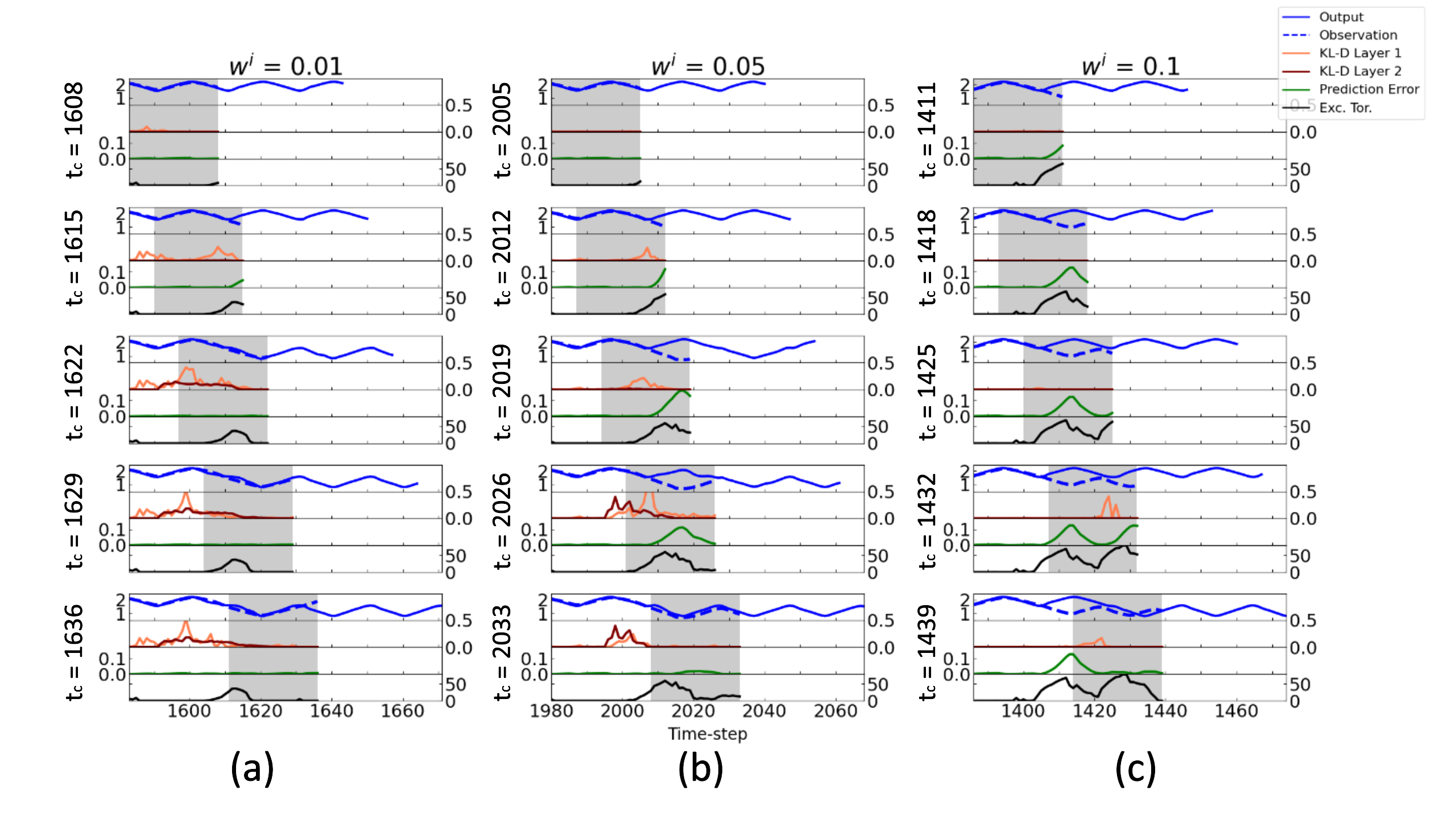}
\caption{Time-development of excess torque, prediction error, and KL-divergence between the approximate posterior and the prior in trained movement transitions
in cases with three different meta-prior $w^i$ settings with $0.01$, $0.05$, and $0.1$. The grey area represents the past window where the head of the window is the current time. The window is shifted 5 times during the movement transition \emph{AB}.
}
\label{fig:TimeDevelopmentDifferentW}
\end{center}
\end{figure*}

First, we examined time-development of essential values during movement pattern transitions induced by the experimenter for each case with a different meta-prior setting.
Fig.\ref{fig:TimeDevelopmentDifferentW} shows an example snapshot of future prediction and past reflection, which shifted every 7 time-steps of the current time during the transition \emph{AB} performed under different settings of meta-prior, $w^i = 0.01, 0.05, 0.1$.
Each snapshot shows one of the observed joint angles $\theta$ (dotted blue) and its prediction $\Bar{\theta}$ (blue) in the top row, the KL-divergences between the approximate posterior and the prior in the layer 1 (orange) and in the layer 2(dark orange) in the second row, the prediction error (green) in the third row, and the excess torque (black) in the bottom row.
The grey area represents the past window where the approximate posterior in terms of adaptive variables $\mathbf{A}_t^{\mu}, \mathbf{A}_t^{\sigma}$ is updated.
We provide two supplementary videos for experiment-1 showing the interaction between the experimenter and Torobo, as well as network dynamics in the case with the meta-prior $w^i$ set to 0.01 (\href{https://youtu.be/jQGnfPMAWes}{video-link1}) and 0.1 (\href{https://youtu.be/th1-5Ay603Y}{video-link2}).

The sequences of time-shifted snapshots in Fig.\ref{fig:TimeDevelopmentDifferentW}, show that excess torque appears first followed by rises in the prediction error (negative log-likelihood) and the KL-divergence. 
Later, the predicted joint angle pattern shifts from pattern \emph{A} to \emph{B} while the gap between the observed joint angle and the reconstructed joint angle remains in the past window.
This is the same for all three cases with different $w^i$ settings.

However, we can see some qualitative differences in the transition process depending on the $w^i$ setting.
The prediction error and the excess torque in the case with a small $w^i$ ($w^i=0.01$) are smaller than those in the case with large $w^i$ ($w^i=0.1$).
Also, the error and torque with a small $w^i$ are less persistent than those with a larger $w^i$.
However, the KL-divergence in the case with a small $w^i$ is larger and persists longer than that in the case with large $w^i$.
In order to confirm these observations, we conducted statistical analysis on the excess torque, KL-divergence, and prediction error time-averaged for each transition period (100 steps). 
This computation was repeated 10 times for each of three different trained networks set with three different $w^i$ values.

\begin{figure}[h] %
\begin{center}
\includegraphics[width=0.95\linewidth]{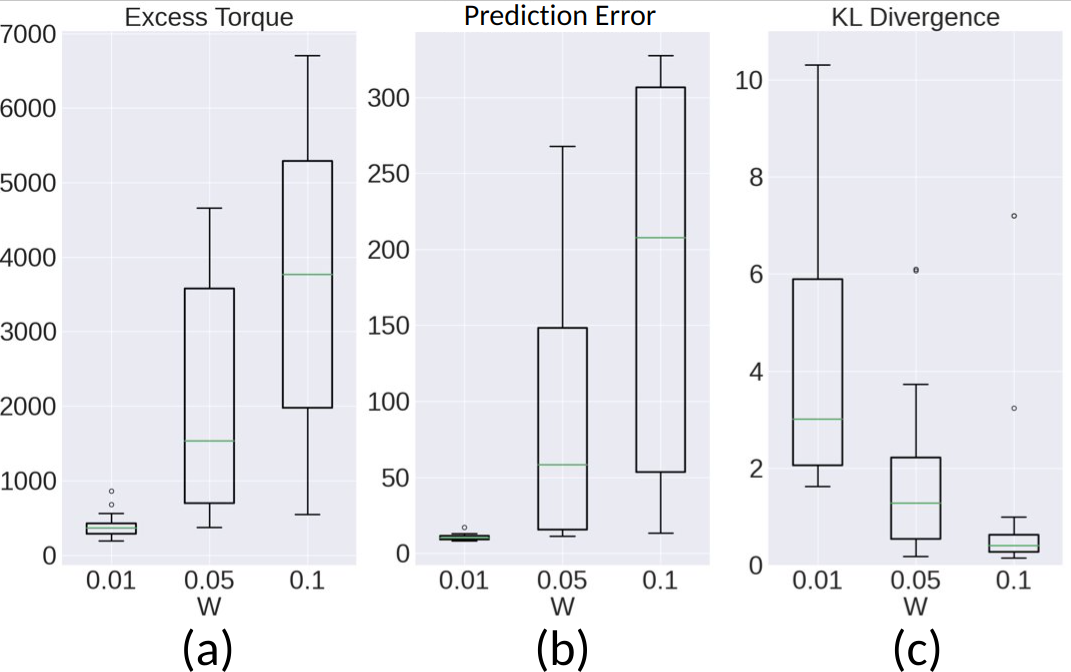}
\caption{Time-averaged excess torque, prediction error, and the KL-divergence between the approximate posterior and the prior in the trained movement transition in cases with three different meta-prior $w^i$ settings with $0.01$, $0.05$, and $0.1$.
}
\label{fig:TransitionForDifferentW}
\end{center}
\end{figure}

The results are shown in Fig.\ref{fig:TransitionForDifferentW}.
%
Both the time-averaged prediction error and the excess torque measured in the cases with small $w^i$ are significantly smaller than those with large $w^i$.
On the other hand, the time-averaged KL-divergence with small $w^i$ is significantly larger than that with large $w^i$.

By considering these statistical results, movement pattern transitions exerted by the experimenter require greater force when $w^i$ is set larger, since the approximate posterior distribution strongly follows the prior distribution representing the current movement intention of PV-RNN, minimizing the KL-divergence between them while the error $\Delta \theta_t$ between the predicted joint angle $\bar{\theta_t}$ and the observed joint angle $\theta_t$ becomes larger.
If the experimenter attempts to move the trajectory of the robot's joint angles in a direction different from that predicted by the PV-RNN, this requires a large excess torque $\tilde{e}_t$ to counteract the large error $\Delta \theta_t$, as derived from Eq. \ref{eq:target joint angle}.

On the other hand, when $w^i$ is set smaller, the approximate posterior follows the prior only weakly, allowing larger KL-divergence between them while the prediction error becomes smaller.
In this case, only a small amount of excess torque is necessary to counteract the small error.
The top-down actional intention of the robot became stronger in the case of larger $w^i$ settings; therefore, the human experimenter was required to exert more force on the robot arms to induce a transition, whereas less force was required with smaller $w^i$ settings, since the top-down actional intention of the robot became weaker.

In the current experiment, the value of $w^i$ was set between $0.01$ and $0.1$.
This is because our preliminary experiments showed that the robot's behaviour became noisy when $w^i$ was set smaller than $0.01$, since the approximate posterior could easily deviate from the prior by noise sampling.
On the other hand, it became difficult for the experimenter to initiate a transition when $w^i$ was set larger than $0.1$, because of substantially increased resistance. 

\subsubsection{Experiment-2}
\begin{figure}[h] %
\begin{center}
\includegraphics[width=0.95\linewidth]{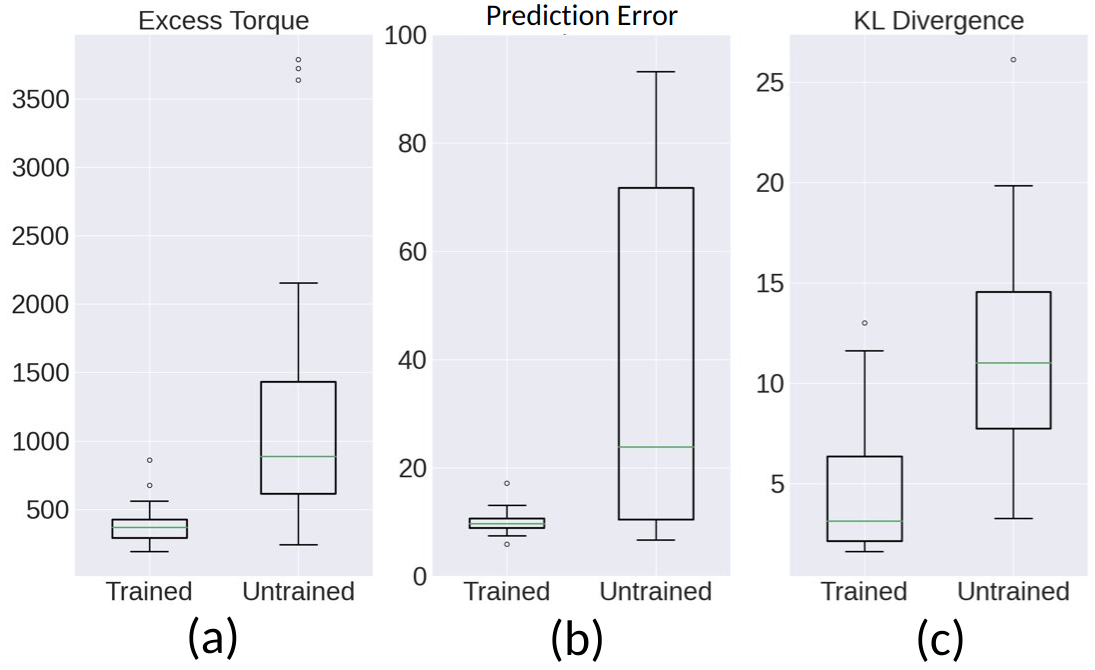}
\caption{The amount of excess torque, prediction error exerted and the KL-divergence between the approximate posterior and the priorc during each attempt for trained and untrained transitions. The meta-prior $w^i$ was set to $0.01$.
}
\label{fig:TrainedVSUntrained}
\end{center}
\end{figure}

Next, we looked at the difference in the excess torque, prediction error, and KL-divergence between trained and untrained transitions while $w^i$ was fixed at a given value.
Both trained and untrained transitions were attempted 10 times for each of the three trained networks with $w^i$ set to $0.01$ for both trained and untrained cases.
However, for the untrained case, among 30 attempts, only 24 succeeded in performing a transition.
Our preliminary experiment showed that the untrained transition became more difficult when $w^i$ was set higher than $0.01$ because of the strong resistance when the robot attempted to lead trained movement transitions.
Fig.\ref{fig:TrainedVSUntrained} shows the time-average of the excess torque, prediction error, and KL-divergence for both trained and untrained transitions attempted by the experimenter.
The time-average of the prediction error, the excess torque, and the KL-divergence are larger in untrained than trained transitions.
This means that untrained transitions require the experimenter to exert more force than for trained transitions because of the free energy, which is the sum of the prediction error and the KL-divergence, and which increases more in untrained transitions.
We provide two supplementary videos for experiment-2 showing the interaction between the experimenter and Torobo as well as the network dynamics in the case with meta-prior $w^i$ set to 0.01, where trained transitions (\href{https://youtu.be/jQGnfPMAWes}{video-link1}) and untrained transitions (\href{https://youtu.be/lZEn5Qvun90}{video-link3}) are performed.


\section{Discussion}
%
%
The current study investigated human-robot bodily interactions via kinaesthesia using a PV-RNN model that was developed based on the free energy principle.
Bodily interactions between a human experimenter and a robot were conducted using Torobo, a humanoid robot equipped with a PV-RNN model that can sense excess torque exerted by a human counterpart.
We especially examined how the counter force between the robot and the human experimenter changed during movement pattern transitions physically guided by the human experimenter, depending on two different condition changes.

In experiment 1, we examined how the setting of a parameter called the meta-prior $w^i$, which regulates the KL-divergence between the approximate posterior distribution and the prior distribution in the interaction phase, affects the counter force generated in executed or attempted transitions.
Results of this experiment showed that in the case of a smaller $w^i$, while the KL-divergence between the approximate posterior and the prior distribution becomes larger, the prediction error (negative log-likelihood) becomes smaller.
Since the prediction error diminishes, the excess torque to counteract this error also decreases.
On the other hand, in the case of a larger $w^i$ setting, while the KL-divergence between the approximate posterior and the prior distribution becomes smaller, the prediction error becomes larger, which requires more excess torque for the transition.
The conflict that appeared between the movement intended by the robot and that executed by the experimenter is distributed to the prediction error and the KL-divergence between the approximate posterior and the prior in proportions determined by the meta-prior $w^i$.
With larger $w^i$ the top-down movement intention of the robot becomes stronger, which results in a stronger counter force, whereas the top-down intention as well as the counter force become weaker with smaller $w^i$.

The above is consistent with past research from our group \cite{wirkuttis2023turn}.
In \cite{wirkuttis2023turn}, Tani conducted simulation studies on synchronized imitative interaction by dyadic, vision-based robots using a PV-RNN model.
That study showed that a robot with smaller/larger $w^i$ tends to follow or lead the other robot set with a larger or smaller $w^i$ with weaker or stronger actional intention in synchronized imitative interaction.
Similarly in the current study, the experimenter easily led
the robot with a smaller $w^i$ with a smaller counter force because of the weaker top-down intention of the robot.
On the other hand, when $w^i$ is quite large, such as $>0.1$ for the robot, it was difficult for the experimenter to lead the robot because of the extremely strong counter force.
In this situation, the experimenter just followed movement patterns strongly led by the robot while grasping the robot's hands, as shown in the preliminary experiment described previously.

In experiment 2, we examined the difference in the counter force required for the experimenter to execute trained and untrained movement pattern transitions.
These experimental results showed that untrained transitions require more force since such transitions are accompanied by larger increases in free energy, the KL-divergence (between the approximate posterior and the prior), and the prediction error.
Trained transitions, on the other hand, require less force because of smaller increases in free energy, the KL-divergence, and the prediction error.

As already mentioned, there have been few studies of human-robot interactions based on the free energy principle.
Although \cite{chameICRA2020} showed that the setting of $w^t$ as meta-prior in the training phase could strongly affect characteristics of human-robot kinaesthetic interactions, the description of their experiment results did not include rigorous analysis with repeated experiments.

The major limitation of the present study is that presented human-robot interactions are not fully interactive since experimenter-induced sequences of movement pattern transitions were determined a priori.
In this regard, Ikegami and his colleagues \cite{ikegami2007turn, iizuka2004adaptability} investigated underlying mechanisms for turn-taking that were generated by spontaneous interaction between artificial agents as well as artificial agents and humans.
Recently, Masumori et al. \cite{Masumori2021} developed a humanoid robot platform, called Alter3, behaviour of which was controlled by sub-modules, including a self-simulator, an automatic mimicry unit, and memory storage, which were perturbed by a specific neurodynamic model for the purpose of conducting experiments on spontaneous human-robot interactions.
They showed that spontaneous turn-taking between imitator and imitated could be developed by autonomous switching of information flow between the two sides.

In future studies, we will undertake human-robot kinaesthetic interaction experiments that assume less a priori.
Such experiments should be done not with experimenters as counterparts of the robots (as in the present study) but by inviting an adequate number of human participants, since the human side also needs to be analyzed.
Such studies will focus on two research issues.
One is to investigate how spontaneous turn-taking can occur in imitative interaction based on kinaesthesis by using the active inference framework \cite{friston2010action, parr2019generalised, baltieri2019pid}.
Spontaneous turn-taking in this setting means that the role of the leader to initiate the next shared patterns switches autonomously between the two sides, such that sometimes the robot may push hard with its own intended patterns and the human counterpart may do so at other times.
This study may require development of an autonomous $w^i$ adaptation scheme, since if $w^i$ on the robot side can shift adaptively by sensing contextual flow in the interaction, the leader-follower relationship should shift accordingly.

The other focus is to investigate how novel movement patterns can be developed through repeated kinaesthetic interaction associated with continuous learning in both robots and human participants, based on the free energy principle.
One assumption is that novel patterns could develop in terms of false memory as the number of movement patterns memorized distributively in the PV-RNN model increases.
This phenomenon of false memory is due to potential non-linearity and stochasticity in the PV-RNN model.
A study on a deterministic RNN model demonstrated this property \cite{tani2004self}.
Novel patterns generated by robots could enhance improvisation of new pattern generation from human counterparts through iterative interaction.

\appendix
\section*{Prior Generation}
\label{app:priorgen}

\begin{figure*}[h] %
\begin{center}
\includegraphics[width=1\linewidth]{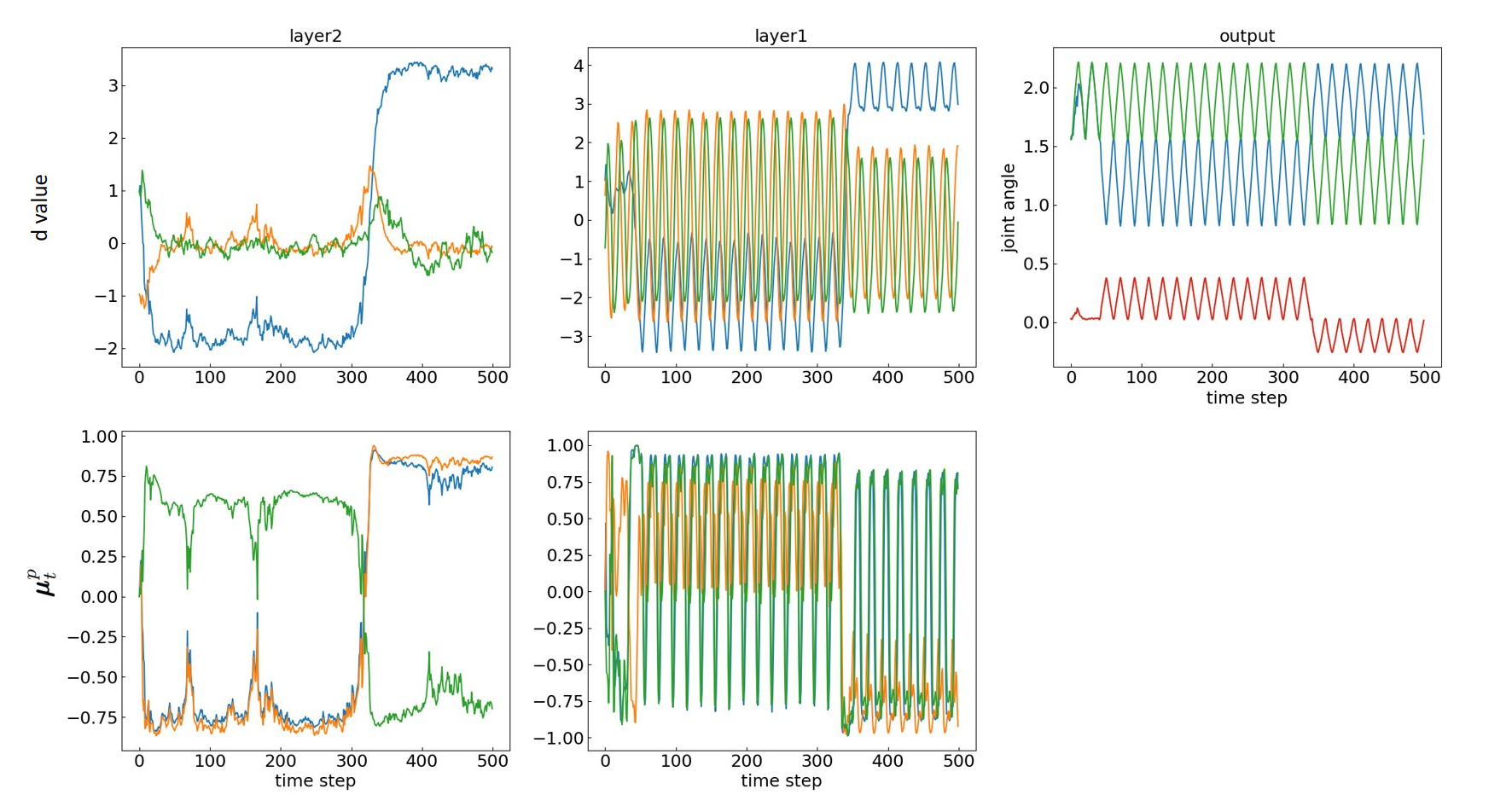}
\caption{Neural activity in prior generation while performing a movement pattern transition. The two columns at the left show neural activities in layers 2 and 1, respectively. The top row shows the 3 principal components of d values, and the bottom row shows the mean value $\mu_t^p$ of the prior. The right column shows the joint angle at each time step. The transition was performed at time step $t = 40, 340$.
}
\label{fig:PriorGeneration}
\end{center}
\end{figure*}

Here, in Fig.\ref{fig:PriorGeneration}, we show part of the result from the prior generation, including two transitions.
As shown in Table \ref{tab:parameters}, since the numbers of d neurons in layers 2 and 1 were 30 and 60, respectively, we performed principle component analysis and reduced both dimensions to 3.

\bibliographystyle{unsrt}  
\bibliography{references.bib}

\end{document}